# 5G Utility Pole Planner Using Google Street View and Mask R-CNN

Yanyu Zhang, Osama Alshaykh

*Abstract*—With the advances of fifth-generation (5G)[1] cellular networks technology, many studies and work have been carried out on how to build 5G networks for smart cities. In the previous research, street lighting poles and smart light poles are capable of being a 5G access point[2]. In order to determine the position of the points, this paper discusses a new way to identify poles based on Mask R-CNN[3], which extends Fast R-CNNs[4] by making it employ recursive Bayesian filtering and perform proposal propagation and reuse. The dataset contains 3,000 high-resolution images from google map. To make training faster, we used a very efficient GPU implementation of the convolution operation. We achieved a train error rate of 7.86% and a test error rate of 32.03%. At last, we used the immune algorithm[5][6] to set 5G poles in the smart cities.

## I. INTRODUCTION

The birth of machine learning marks the breakthrough of the human dream of intelligent machines[7]. Current approaches to object recognition make essential use of machine methods. Nowadays, machine learning is used everywhere in the world.

According to media reports, 5G technology is receiving worldwide attention until now. It is reported that the EU 5G Infrastructure Public Private Partnership(PPP) is expected to start 5G technology trials in 2018, and Japan plans to achieve 5G commercialization before the 2020 Tokyo Olympics. In addition, South Korea is also likely to conduct 5G pre-commercial trials in early 2018 and achieve 5G commercialization by the end of 2020.

5G will quite literally reshape the high-speed broadband service industry for consumers and the enterprise. It will also help enable technology innovation that will require ultra-high capacity, low latency networks to be fully realized, for example, smart city applications and autonomous vehicles. When you think of the city of the future, with a significant number of autonomous cars driving around, there are a number of wireless solutions that will need to be deployed to allow this futuristic vision to become a reality.

In order to help company to install 5G equipment on utility poles, we build a dataset from Google Static Street View and labeled by VGG Image Annotator(VIA)[8]. Then training the dataset base on Mask R-CNN structure with an efficient NVIDIA GeForce RTX 2080 Super GPU. After training, we got a train test error rate of 7.86% and a test error rate of 32.03%. Then we downloaded all of the street view pictures in Boston from Google Map and detect whether there are poles in these pictures or not. After that, we marked poles as points on the map, and use immune algorithm to decide the best and least number of poles that will be built in Boston.

In the final, we use Django to make a web page and all resources have been open-sourced on GitHub(https://github.com/zhangyanyu0722/5G-Utility-Pole-Planner).

## II. THEORY AND APPROACHES

### A. Model Training

In this paper, we used an efficient approach to detect objects in the images, called Mask R-CNN, which outperforms all existing, single-model entries on every task, including the COCO 2016 challenge winners[9].

Mask R-CNN is a development of Faster R-CNN by adding a branch for predicting segmentation masks on each Region of Interest (RoI)[10], in parallel with the existing branch for classification and bounding box regression. (Figure 1)

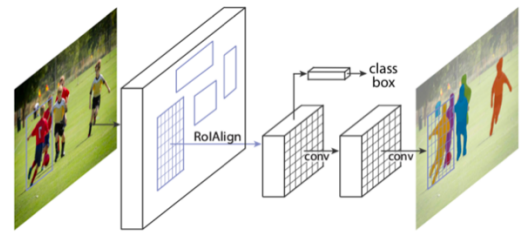

Figure 1. The Mask R-CNN framework for instance segmentation.

We retrained our new model based on the old weight file: mask_rcnn_coco.h5. Our aim is to divide whether there are poles exiting in pictures or not. So we set two classes for our dataset: poles and no-poles. And we trained for 25 epochs with one GPU. After about one days training, we got the best

train test error rate of 7.86% and test error rate of 32.03%. (Figure 2)

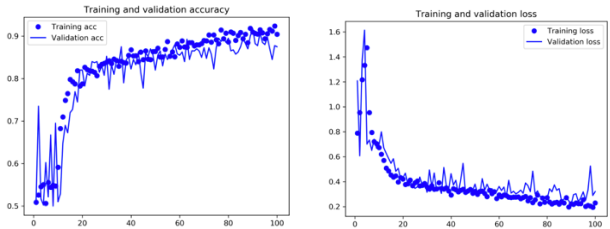

Figure 2. The Mask R-CNN training and validation accuracy and loss.

B. Model Quality

In order to observe the effect of the model, we printed 4 origin images and 4 predicted image through our model randomly. (Figure 3) We can clearly observe the second and fourth predicted images each include two poles, different to the origin images. The third predicted image detect a piece of wire as a pole. But overall our model works very well.

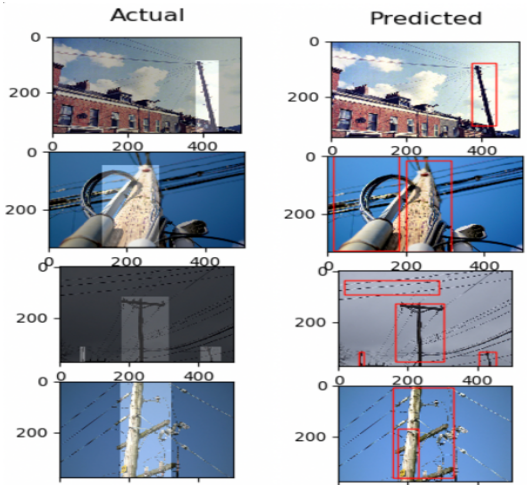

Figure 3. The comparison of actual and predicted images

C. Immune Algorithm

In artificial intelligence[11], artificial immune systems (AIS)[12][13][14] are a class of computationally intelligent, rule-based machine learning systems inspired by the principles and processes of the vertebrate immune system. Artificial Immune Systems (AIS) are adaptive systems, inspired by theoretical immunology and observed immune functions, principles and models, which are applied to problem solving.

The first step of immune algorithm is to identify antigen. The antigen is objective function and various constraints. Then the initial antibody will be randomly generated. The key is calculating the affinity, which will feedback a fitness value. If this fitness key is satisfy the break condition, it will output the result. If not, immune algorithm will do immune processing include immune selection, cloning, mutation, and suppression. After that, the new antibody set will be produced and refresh all the cell in the initial antibody. (Figure 4)

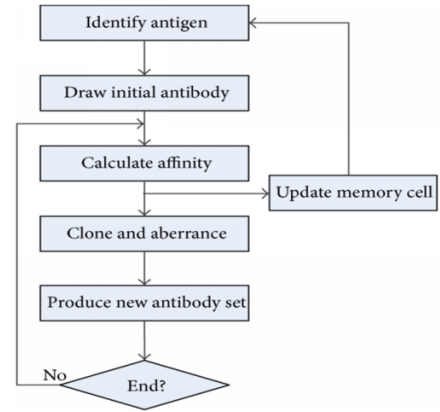

Figure 4. Immune Algorithm Flowchart

III. EXPERIMENTS AND RESULTS

In order to test our 5G utility poles planner tool simply and clearly, we build a webpage on Django[15]. (Figure 5) After input the latitude and longitude of the left-top and right-down position in a rectangle. Then press the "submit" button showing below.

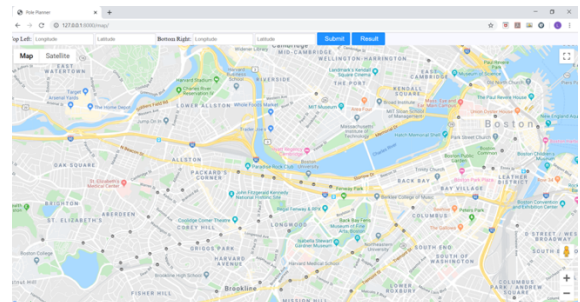

Figure 5. Django webpage

The background will download all the images as an initial step, then detect the poles in the images with the help of our prepared model. After that, the immune algorithm will finish the iteration and show the result when you press the "Result" button. (Figure 6)

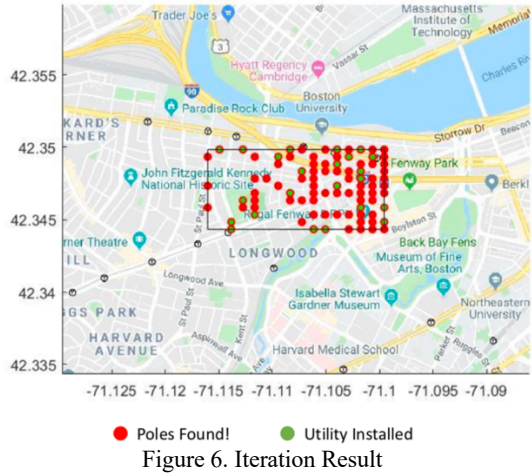

Figure 6. Iteration Result

In the picture(Figure 6), the red points show all of the utility in this region, while green points show the final decided points to use, which 5G signal can cover all the areas.

## IV. DISCUSSION

During the test process, we tried twenty different areas in MA, also concludes some regions crossing the river. For some regions, we found some "error points" (Figure 7) Clearly there are two points in the river. The reason of these error points is the quality of the model. Although these two points has not been used to poles installed, they should be delated.

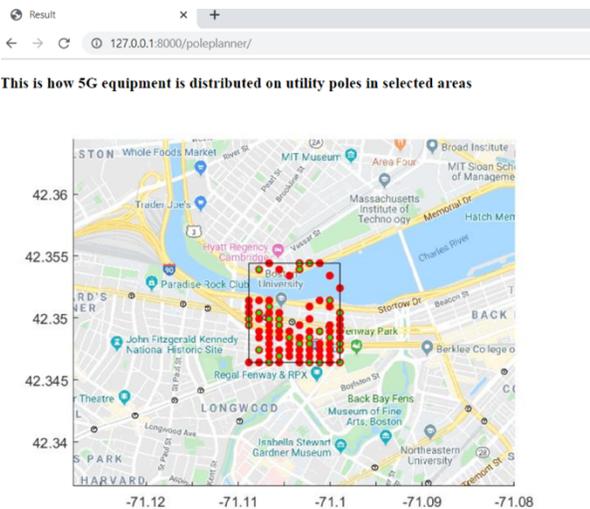

Figure 7. Error Points

On the other hand, this 5G poles planner system need a long time to run out the results. We divided the whole process into three parts: download images from Google Map, detect poles in the images and immune algorithm to plan. In the final, we found most of the time speed on immune algorithm, because it needs long time to iteration.

## V. CONCLUSION AND FUTURE WORK

In this study, we designed a 5G utility poles building system, which can help the company to install 5G equipment in the specified region. Our system used 3,000 images from Google map and trained by a recently new method named Mark R-CNN. In the final, we plan the poles by the immune algorithm and show the result in Django.

In the future, we need to improve our model by increasing the dataset. Also, it is not easy to find the accuracy position by observing such a result. We will build another UI to make planning simple. What is more, the system spends a large amount of time on iteration during the immune algorithm. We will pay more attention to reduce the run time. With the development of GPU and algorithm, the 5G utility pole planner will become faster and useful.